\pdfoutput=1

\documentclass[11pt]{article}

\usepackage[preprint]{acl}

\usepackage{times}
\usepackage{latexsym}

\usepackage[T1]{fontenc}

\usepackage[utf8]{inputenc}

\usepackage{microtype}

\usepackage{inconsolata}
\usepackage{booktabs}
\usepackage{soul}
\usepackage{color}

\usepackage{multirow}
\usepackage{multicol}
\usepackage{enumitem,kantlipsum}
\usepackage[normalem]{ulem}
\usepackage{color,colortbl}
\usepackage{todonotes} 
\usepackage{soul}
\usepackage{threeparttable, makecell}
\usepackage{makecell}
\usepackage{xcolor}

\usepackage{tcolorbox}

\usepackage{comment}
\usepackage{array}  
\usepackage{lscape}  
\usepackage{multirow}
\usepackage{graphicx}
\usepackage[normalem]{ulem}
\useunder{\uline}{\ul}{}

\usepackage{array}  
\usepackage{multirow}  
\usepackage{geometry}  
\usepackage{algorithm}
\usepackage{algpseudocode}
\usepackage{amsmath}
\usepackage{subcaption}
\usepackage{fancybox}
\usepackage{tikz}
\usepackage{listings}
\usepackage{color}

\usepackage{pifont}

\usepackage{enumitem}
\setlist{nosep,topsep=-\parskip}

\usepackage{soul}
\usepackage{xcolor}
\definecolor{dkgreen}{rgb}{0,0.6,0}

\usepackage{arabtex}
\usepackage{utf8}
\setcode{utf8}

\usepackage{graphicx}
\definecolor{lightblue}{rgb}{.50,.95,1}
\definecolor{tri}{rgb}{.25,.88,.82}
\definecolor{lilac}{rgb}{0.85,0.64,0.85}

\newcommand{\nqa}{\textbf{\emph{NativQA}}}
\newcommand{\nqaf}{\textbf{\emph{NativQA framework}}}

\newcommand{\code}[1]{\texttt{#1}}
\usepackage{amssymb}
\usepackage{hyperref}
\usepackage{verbatim}
\usepackage{setspace}

\usepackage{colortbl}
\usepackage{booktabs}
\usepackage{cuted}
\title{NativQA Framework: \\Enabling LLMs and VLMs with Native, Local, and Everyday Knowledge} 

\author{
    Firoj Alam,\textsuperscript{$^1$}
    Md Arid Hasan,\textsuperscript{$^2$}
    Sahinur Rahman Laskar,\textsuperscript{$^3$}\\
    {\bf
    Mucahid Kutlu,\textsuperscript{$^4$} 
    Kareem Darwish,\textsuperscript{$^1$}
    Shammur Absar Chowdhury\textsuperscript{$^1$}    
    } \\
    \textsuperscript{$^1$}Qatar Computing Research Institute, Qatar, \\      
    \textsuperscript{$^2$}University of Toronto, Canada, 
    \textsuperscript{$^3$}UPES, India, 
    \textsuperscript{$^4$}Qatar University, Qatar \\   
    fialam@hbku.edu.qa, arid@cs.toronto.edu  \\
    \url{https://gitlab.com/nativqa/nativqa-framework}
}

\begin{document}
\maketitle

\begin{strip}
\vspace{-6mm} 
\centering
\includegraphics[width=\textwidth,height=0.26\textheight,keepaspectratio]{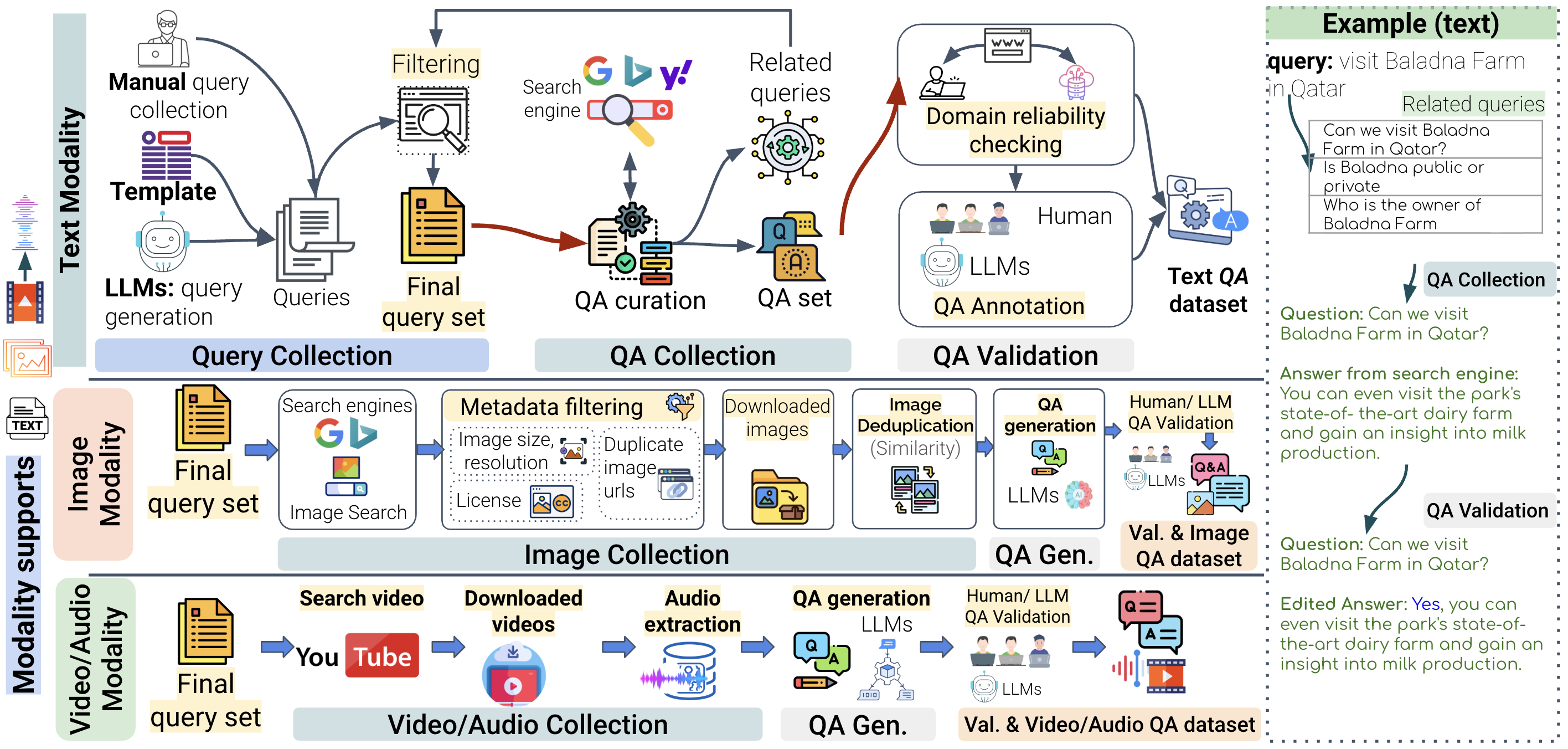}
\captionsetup{type=figure,skip=25pt}
\caption{\nqaf{}, demonstrating the entire pipeline, from query collection to the final dataset development process.}
\label{fig:nativqa_framework}
\end{strip}

\begin{abstract}
The rapid progress of large language models (LLMs) raises concerns about cultural bias, fairness, and performance in diverse languages and underrepresented regions. Addressing these gaps requires large-scale resources grounded in multilingual, local, and cultural contexts. We systematize and \textbf{extend the earlier \textit{NativQA} framework to multimodality} by adding image, audio, and video support, enabling scalable construction of culturally and regionally aligned QA datasets in native languages. Given user-defined seed queries, the framework uses search engines to collect location-specific everyday information. We evaluate it across 39 locations in 24 countries and 7 languages, spanning extremely low-resource to high-resource settings, and collect over $\sim$300K text QA pairs, $\sim$312K images, and $\sim$29K videos with associated audio. The developed resources can be used for LLMs benchmarking and further fine-tuning. The framework has been made publicly available for the community. Demo video is available here: \href{https://shorturl.at/DAVn9}{https://shorturl.at/DAVn9}.
\end{list}
\end{abstract}




\section{Introduction}
\label{sec:introduction}

Large Language Models (LLMs) have exhibited cultural bias toward certain cultures, largely due to the predominance of training data in high-resource languages and the limited digital representation of low-resource languages \cite{li2024culturellm,pawar2024survey,durmus2024towards}. Given that most LLMs are trained primarily on English or translated data, which often fails to capture how people naturally speak, ask questions, and express their needs in their native languages and dialects. Since the emergence of LLMs, there has been growing concern about how to develop and benchmark models that perform equitably across diverse languages and cultures~\cite{NEURIPS2024_77f089cd,myung2024blend,kim-etal-2024-click,pawar2024survey}. In addition to cultural representation,\footnote{Cultural ability includes understanding the diversity of cultural elements across different cultures \cite{pawar2024survey}.} there is also concern about whether LLMs can effectively respond to everyday, long-form, and complex queries that require paragraph-length answers \cite{arora2025calmqa,myung2024blend}. These concerns have driven efforts to develop multilingual~\cite{ustun2024aya,team2024gemma} and language-specific~\cite{fanar2024,nahin2025titullms,nguyen-etal-2024-seallms} closed and open-source LLMs.




To benchmark and fine-tune models with region- and culture-specific knowledge, prior work has developed language- and culture-specific resources and corresponding benchmarks and models \cite{chiu2024culturalbench,shi-etal-2024-culturebank,li2024culturellm}. Common approaches include mining QA from community forums \cite{shi-etal-2024-culturebank}, having native speakers write QA pairs \cite{chiu2024culturalbench}, translating English content into local languages \cite{myung2024blend}, and using LLMs to generate QA pairs \cite{putri2024can}. In multimodality, related efforts include CVQA \cite{romero2025cvqa}, PALO \citep{Rasheed2025PALO}, Pangea \citep{yue2024pangea}, and SPEECH-COCO \cite{havard2017speech}. 

A consistent finding is that most LLMs perform well on high-resource languages but underperform in mid- and low-resource settings \cite{myung2024blend}. Fine-tuning with culture-specific knowledge can also improve performance \cite{li2024culturellm,NEURIPS2024_77f089cd}. These results motivate datasets that better represent diverse regions, cultures, and native content. However, building such resources remains costly and labor-intensive, and scalable dataset creation is still limited due to the lack of tools.

To address these gaps, \citet{hasan2024nativqa} introduced the initial \nqa{} pipeline and applied it to build \textsc{MultiNativQA}, a $\sim$64K QA dataset in 7 languages across 18 topics. \textbf{This demo paper has a different focus: we redesign and generalize the pipeline} by \textit{(i)} consolidating the earlier scripts into a reusable, configurable framework; \textit{(ii)} \textbf{extending it with multimodality (image, video, and extract audio from the video)} QA collection and \textbf{pluggable search backends}; and \textit{(iii)} providing practical guidelines and cost estimates for running the framework at scale. In this sense, \textsc{MultiNativQA} is one use-case of the broader framework released here. \textit{Chronologically}, the \textsc{MultiNativQA} paper appeared first and outlined a generalized framework as future work. \textbf{This demo paper delivers that extended framework with new capabilities} and broader case studies.

\noindent Our contributions in this paper are as follows:
\begin{itemize}[noitemsep,topsep=0pt,leftmargin=*,labelsep=.5em] 
    \item We systematize and extend the earlier \nqaf{} pipeline into a modular, reusable, open-source framework. 
    \item The framework offers several key features:  
    \textit{(i)} seamless integration of both user- and LLM-generated queries to collect QA pairs;  
    \textit{(ii)} location-agnostic -- users can collect data from any region;  
    \textit{(iii)} support for multiple search engines;  
    \textit{(iv)} multimodal (e.g., image, video) search capability;  
    \textit{(v)} automated removal of duplicate QA pairs; 
    \textit{(vi)} domain reliability checks; and
    \textit{(vii)} an efficient caching mechanism to reduce redundant API calls, leading to cost savings and mitigating time-out issues.
    \item We evaluated it for 39 locations, from 24 countries covering 7 languages and 18 topics, which resulted over $300K$ data points for text, $\sim$312K for images, $\sim$29K for videos/audios. 
    Unlike \citet{hasan2024nativqa}, which only considers text-based QA pairs, this implementation also supports collecting image, video and extracted audio.
    \item We report implementation details (e.g., search‑engine backend, caching strategies) and cost estimates for running the pipeline, to help practitioners budget and adapt NativQA to their own settings.
    \item We have made the \nqaf{} publicly available for the community.    
\end{itemize}

\section{\nqaf{}}
\label{sec:nativqa_framework}


In Figure \ref{fig:nativqa_framework}, we present the \nqaf{} consisting of three modalities: \textit{(i)} Text, \textit{(ii)} Image, and \textit{(iii)} Video/Audio modalities. The left part of the figure illustrates the entire process, while the right part shows an example of the query collection to QA validation process for the text modality. In the following section, we discuss them in detail. 

\subsection{Text Modality}
Text modality consists of three interconnected modules, such as \textit{(i)} Query Collection, \textit{(ii)} QA Collection, and \textit{(iii)} QA Validation,  discussed below.

\subsubsection{Query Collection (QC)}
The objective of this module is to collect open-ended queries focused on various \textit{predetermined topics} derived from common knowledge in everyday communication. The set of topics should be established prior to beginning the query collection process, as this facilitates the gathering of topic-specific queries. 
The framework has been evaluated using a predetermined set of 18 topics that are culture- or region-dependent (e.g., events, literature, etc.). However, the topics can be anything and can be task specific. The query collection process can be carried out using three different approaches: 
\begin{itemize}[noitemsep,topsep=0pt,leftmargin=*,labelsep=.5em]
    \item \textbf{Manual query collection:} Involves asking annotators to write topic-specific queries they might pose when seeking everyday information.
    \item \textbf{Template-based generation:} Defines templates for different topics with a placeholder for the location. For example, \textit{``main cultural festivals in [LOCATION]''}, where LOCATION can be a city, country, or region name.
    \item \textbf{LLM-based generation:} Uses LLMs to generate queries.
\end{itemize}


The manual query collection process requires recruiting location- or region-specific annotators. Moreover, like any other annotation task, it is time-consuming.
In contrast, the latter approaches are more cost- and time-effective.
The template-based approach involves defining location-agnostic templates, which can be seamlessly adapted for different locations. The LLM-based approach requires prompting to generate appropriate queries. We use the term \textit{seed query} ($Q_0$) to refer to the queries generated by all approaches. These queries are then used in later steps by the QA collection module to gather QA pairs and additional queries. 

\noindent \textbf{Filtering:} Given that queries can have exact and/or near duplicates, it is important to remove such duplicates as part of the filtering process. The \textit{filtering} module handles this by using exact matching for exact duplicates and a similarity-based approach for near duplicates. These steps results in the \textit{\textbf{final set of queries}}. 
 
\subsubsection{QA Collection (QAC)}
The next step is to collect QA pairs using a search engine, e.g., Google. Most search engines support features like ``People also ask'' or ``Related queries,'' which list questions asked by real users that are potentially relevant to the initial seed query, as shown in Figure~\ref{fig:google_example}. Moreover, these questions are accompanied by answers extracted by the search engine, along with links to the sources of the answers.


The \code{QA collection module} implements Algorithm~\ref{alg:collect_qa_pairs} (Appendix). It takes the seed queries $Q_0$ and the number of iterations $N_{\text{iter}}$ as input. For each iteration $i \in N_{\text{iter}}$, it collects QA pairs $P_{QA}^i$ and related queries $S^i_{\text{rel}}$ for each query $q \in Q$, then passes them to the \code{filtering module} and updates the current query set $Q$. This process is repeated for all iterations to obtain the \code{final QA set}, $S_{QA}$, for the enriched queries $Q$. The idea of the \code{filtering module} is to remove duplicates that may arise in different iterations. 

We collect location- and language-specific QA pairs by querying search-engine APIs with explicit country, language, and city-level parameters. Our implementation uses SerpAPI\footnote{\url{https://serpapi.com/}} to access structured widgets (e.g., ``People also ask'' and related questions), and stores extracted questions, answers, and source URLs as candidate QA pairs.

\subsubsection{QA Validation (QAV) Module}
The next step of the framework is to validate the QA pairs. It is important for several reasons: \textit{(i)} the retrieved related queries might be less relevant, \textit{(ii)} answers might be incomplete, \textit{(iii)} the source of the answer can be less reliable. For the \code{QAV} the framework integrates the below two approaches.

\noindent \textbf{Domain Reliability Checking (DRC).} 
The answers collected by \nqaf{} include source URLs. The DRC module retains QA pairs by filtering for reliable domains, under the assumption that answers from such sources are more likely to be trustworthy and suitable for inclusion. The framework currently supports two DRC options:
\begin{itemize}[noitemsep,topsep=0pt,leftmargin=*,labelsep=.5em]
    \item Based on the domains of the QA pairs listed in $S_{QA}$, annotators manually check and label them as \textit{very reliable}, \textit{partially reliable}, \textit{not sure}, or \textit{completely unreliable}. 
    For the annotation task, the annotation guidelines proposed by \citet{hasan2024nativqa} can be used. 
    \item In the current implementation, \nqaf{} provides $2,080$ source URLs curated from the \textit{MultiNativQA} dataset, where annotators identified them as reliable. These domain URLs can be used to verify the reliability of the answers.
\end{itemize}
This approach is more practical and scalable for large-scale QA pairs, as it reduces the manual effort needed to obtain hand-curated answers.


\noindent \textbf{QA Annotation (QAA).} 
To increase the reliability of the curated QA pairs, different steps can be taken. Currently \nqaf{} supports two different approaches: 
\begin{itemize}[noitemsep,topsep=0pt,leftmargin=*,labelsep=.5em]
\item \textbf{Manual annotation:} This approach involves manually reviewing and editing the answers, which includes three types of annotations.
\textit{(i)} \textit{Question validation:} Annotators label each question as \textit{good} or \textit{bad}. A \textit{good} question is fact-seeking and can be answered with an entity or an explanation. A \textit{bad} question is ambiguous, incomprehensible, based on a clear false presupposition, opinion-seeking, or otherwise not factual.
\textit{(ii)} \textit{Answer editing:} If the answer is incomplete, contains extra details, or includes incorrect information, annotators edit it to be accurate and complete. To keep the process grounded, annotators must rely only on the provided source web page.
\textit{(iii)} \textit{Location relevance:} Annotators determine whether the question is relevant to the specified \mbox{[LOCATION]}.
\item \textbf{LLM-based annotation:} While manual annotation is more reliable, it is also time-consuming and costly. To reduce such effort, \nqaf{} supports the use of LLM-based approaches as models like GPT-4 have demonstrated effectiveness in improving task-specific model performance~\cite{li-etal-2023-synthetic,li2024data}. 
In the LLM-based approach, a concise version of the annotation guidelines \cite{hasan2024nativqa} 
is adapted into the prompt to check the question, edit the answer, and identify location relevance.

\end{itemize}

\subsection{Image Modality}
\label{sec:image_modality}

It consists of three interconnected modules: \textit{(i)} Image Collection, \textit{(ii)} QA Generation, and \textit{(iii)} Image QA Validation, discussed below.


\noindent
\textbf{Image collection.}
This module collects images from search engines (e.g., Google and Bing). Like the text pipeline, \code{Image Collection} queries search-engine APIs with explicit country, language, and city-level parameters using a set of seed queries. Note that the query generation process is the same across all modalities; we refer to the resulting set as the \textit{\textbf{Final query set}}. We interface with multiple engines using SerpAPI, which provides structured outputs such as image results, suggested searches, and related-search widgets. Retrieved images are added to the candidate pool, while suggested and related queries are used to expand coverage in subsequent iterations. For each candidate, we store the input query, thumbnail URL, image URL, and metadata (size and resolution), and filter duplicate URLs during collection. We then download the candidates and remove near-duplicate images using an embedding-based similarity method following \cite{alam-etal-2024-armeme}.

\noindent
\textbf{QA generation.}
Given that developing robust multimodal and multilingual QAs is challenging due to the significant time and cost involved in creating them manually \citep{changpinyo-etal-2023-maxm}. In contrast, recent advances in LLMs and VLMs enable scalable generation of diverse QA pairs~\citep{zhang2025automated}. Therefore, for open-ended QA generation we use VLMs (GPT-4.1).

\noindent
\textbf{Image QA validation.}
Prior work shows that strong LLM judges can closely match human preferences at scale, and recent VLM judges extend this idea to fine-grained multimodal evaluation \citep{zheng2023judging,lee-etal-2024-prometheus}. Hence, to scale the annotation/validation process, we use VLMs for QA validation, whether answer is (\textit{i}) clear and informative, and (\textit{ii}) plausibility \& faithfulness. \textit{Human in-the-loop will be more reasonable approach to balance reliability, scale and cost}.  

\subsection{Video/Audio Modality}
The video/audio modality consists of three interconnected modules: \textit{(i)} Video/Audio Collection, \textit{(ii)} QA Generation, and \textit{(iii)} Video/Audio QA Validation. We discuss them below.


\noindent
\textbf{Video/Audio Collection.}
This module collects videos from video search engines such as YouTube. Similar to the text and image pipelines, \code{Video/Audio Collection} queries search APIs with explicit country, language, and, when supported, city-level parameters using a set of seed queries. The APIs return structured metadata such as titles, descriptions, thumbnails, durations, key moments, and media URLs, along with suggested and related queries. Retrieved items are stored as candidates while suggested and related queries are iteratively added to expand coverage. We filter duplicate URLs during collection, then download the remaining candidates in parallel as video and audio streams using asynchronous workers.

Note that we do not explicitly search for audio. Instead, we extract the audio track from each retrieved video, a common practice in large-scale audio and speech dataset construction from YouTube, including AudioSet and VGGSound \citep{gemmeke2017audioset,chen2020vggsound}. 


\noindent
\textbf{Video/Audio QA Generation \& Validation.}
We generate QA pairs for video and extracted-audio content using VLMs, and annotation/validate them using a VLMs, following the same approach as in image QA generation and validation.

\subsection{Executing the Framework}
\label{sec:framework:running}


Once the seed queries are curated, running \nqaf{} is straightforward and requires a single command with configurable parameters. It reads \textit{seed queries} from an input file (\code{--input\_file <$file$>}) and, for each query, calls the search API with parameters such as \code{--country\_code <$c$>}, \code{--location <$l$>}, and the iteration limit $n$ (\code{--limit <$n$>}). We provide an example command below. The framework also includes additional utilities such as domain reliability checking, described in the framework's documentation.

\begin{lstlisting}[language=bash,basicstyle=\ttfamily\small]
  $python -m nativqa --engine google --search_type text --input_file <data/test_query.csv> --country_code qa --location <"Doha, Qatar"> --env <envs/api_key.env> --n_iter n
\end{lstlisting}
\vspace{-0.3cm}
\section{Evaluation of \nqaf{}}
\label{sec:experiments}

\noindent\textbf{Evaluated in multiple studies.} 
While the MultiNativQA dataset development is the primary case study, \nqaf{} has also been used in several other works. \citet{mousi-etal-2025-aradice} employed the framework to collect Arabic cultural and dialectal QA data for benchmarking LLMs, and \citet{hassan-bhatti-etal-2025-cultranai} used it to curate Arabic culturally specific QAs across different locations. 

\noindent\textbf{Evaluation through additional data collection.} We further evaluated the framework for additional 30 locations by creating seed query templates. We developed Arabic and English seed query templates from the seed queries that were used to collect the \textit{MultiNativQA dataset}. 
We selected seed queries that contain mentions of locations in the \textit{MultiNativQA}, which were then used with the \textit{[LOCATION]} template (e.g., \textcolor{blue}{\textit{women's wedding attire in [LOCATION]}}). The \textit{[LOCATION]} template was subsequently replaced with other locations to create seed queries for different regions. We manually validated the seed queries generated from the template and removed those that were specific to a particular location (e.g., \textcolor{blue}{\textit{1971 liberation war in Bangladesh/[LOCATION]}}). For Arabic, we used only the seed queries from the template that were employed to collect QA data for Qatar, while the English queries were adopted from the manual seed queries for both Bangladesh and New York.
In Appendix~\ref{sec:app_additional_dataset}, we discuss the details of the text-based QA data distribution (over 300K) that have been collected using the \nqaf{}. 

\noindent \textbf{Leveraging LLMs.} The \textit{framework} uses LLMs/VLMs at different stages, query generation, QA generation across modalities, and LLM-based QA annotation/validation. These steps are implemented through prompting, and we release the corresponding prompts and scripts as part of the framework.


\noindent \textbf{LLMs-based Query Generation.} We generate queries by prompting an LLM with the target location and topic. To assess quality, we manually reviewed a random sample and found the queries acceptable to annotators, aligned with human-written queries, and consistent with the specified location and topic.



\noindent \textbf{VLMs QA Generation.}
For QA generation from images, videos and audio we use VLMs, which has also been commonly used in the current state-of-art as mentioned earlier.  

\noindent \textbf{LLMs/VLMs-based QA annotation/validation.} We adapt a concise version of the annotation guidelines into the prompt so the LLM can validate questions, edit answers, and assess location relevance. To manually assess annotation quality, we use two settings, \textit{(i)} QA validation by selecting the best answer and \textit{(ii)} QA annotation on a sampled subset of LLM-labeled pairs.

\begin{itemize}[noitemsep,topsep=0pt,leftmargin=*,labelsep=.5em]
\item \textbf{Manual QA validation selecting the best answer.} For this evaluation, we manually evaluated 500 text based QA pairs from Egypt, which is due to the availability of Egypt-based annotators. The goal was to compare the LLM-edited answer against the original answer retrieved from the search engine. Annotators selected the better answer based on accuracy, clarity and readability, and completeness and relevance (Appendix~\ref{app:qa_validation}). They could choose \textit{(i)} \textit{Answer 1} (original), \textit{(ii)} \textit{LLM-edited answer}, or \textit{(iii)} \textit{neither}. Two native Arabic-speaking annotators residing in Egypt completed the task and were compensated at a standard third-party rate. We report inter-annotator agreement using observed agreement, Gwet's AC1~\cite{gwet2008computing}, and Cohen's $\kappa$. Observed agreement and Gwet's AC1 were 0.842 and 0.806, indicating strong alignment. The high observed agreement suggests that using LLMs to edit answers may be a promising direction.

    
\item \textbf{Manual QA validation answer scoring.}
We manually assess the quality of LLM-edited answers (text-based) by assigning Likert-scale scores from 1 to 5 for \textit{clarity}, \textit{faithfulness}, \textit{informativeness}, and \textit{plausibility}. We adopt these criteria from prior work on evaluating natural language explanations \cite{huang-etal-2024-chatgpt,zavolokina2024think}, and provide detailed guidelines in Appendix~\ref{app:verifying_quality}. Based on annotator availability, we evaluate two locations, Sudan and Yemen, using approximately 350 examples per test set. Each answer is scored by three annotators for Sudan and two annotators for Yemen.
We report average Likert-scale scores for each metric in Table~\ref{tab:annotator_likert_score}. We also measure agreement on ordinal ratings using $r^*_{wg(j)}$ \cite{james1984estimating}, which compares the observed rating variance to the maximum variance under complete disagreement. Scores of 0.85 for Sudan and 0.86 for Yemen indicate strong agreement~\cite{o2017overview} (Table~\ref{tab:annotator_likert_score}). 
Overall, these results suggest that LLMs can effectively support QA annotation.

\end{itemize}

\noindent
For query generation, QA generation, and LLM-based QA annotation, we used GPT-4o (version 2024-11-20) and Gemini 2.5 Pro. However, the framework is model-agnostic and can use any suitable LLM or VLM for these tasks.

\noindent
\paragraph{Cost Estimation and Comparison.}
The main recurring cost of running NativQA at scale is the search-engine API. For each seed query, the QA Collection module issues one request and extracts multiple related questions and answers from the ``People also ask'' widget and related-query suggestions. Building our 300K QA pairs cost approximately \$2,750 in total, including API usage and human validation on a small sample, which is about \$0.009 per QA pair. Human-only pipelines are far more expensive. CaLMQA~\cite{arora2025calmqa} pays \$1.5 per question for high and mid-resource languages and \$1.8 per QA for low-resource languages.

\section{Features of \nqaf{}}
\label{sec:features}
\nqaf{} is a generic framework that is agnostic to location and language. It supports scalable regional, multimodal and multilingual data collection, while remaining simple to run and flexible to customize.

\noindent\textbf{Modularity.}
As shown in Figure~\ref{fig:nativqa_framework}, \nqaf{} follows a loosely coupled design that separates seed query collection, QA curation, and QA validation. Their interactions support a modular and extensible architecture.

\noindent\textbf{Generality.}
Users can configure the location, language, search engine, and modality, with support for text and image search. All outputs and metadata are stored in JSON and JSONL, which simplifies downstream use and adaptation.

\noindent\textbf{Caching.}
API timeouts and failures can force costly re-runs. \nqaf{} mitigates this with caching of intermediate outputs per sample so re-runs reuse cached results and avoid repeated API calls, reducing cost and effort.

\noindent\textbf{Open-source.}
We release \nqaf{} as open-source to support community. 

\section{Related Work}
\label{sec:related work}

LLMs perform strongly on standard NLP benchmarks \cite{ahuja2023mega,hendy2023good,abdelali-etal-2024-larabench}, yet they often struggle with culturally grounded tasks, especially in low-resource languages and dialects \cite{pawar2024survey}. Recent work therefore develops culturally grounded datasets to benchmark and fine-tune models across languages and regions \cite{arora2025calmqa,shi-etal-2024-culturebank,NEURIPS2024_77f089cd}. Complementary resources target everyday information seeking, including Wikipedia-based QA \cite{kwiatkowski-etal-2019-natural}, Google Search QA \cite{khashabi-etal-2021-gooaq-open}, Reddit \cite{fan-etal-2019-eli5}, and native-speaker QA writing \cite{clark-etal-2020-tydi}. Other approaches combine native and machine-translated data \cite{arora2025calmqa} or leverage LLM-generated QA \cite{NEURIPS2024_77f089cd}. Overall, these efforts highlight the need for models and datasets that reflect diverse linguistic forms, social norms, and local experiences \cite{chiu2024culturalbench}.

Prior work expands cultural and linguistic coverage with datasets such as TyDi QA \cite{clark-etal-2020-tydi}, BLEnD \cite{myung2024blend}, and CaLMQA \cite{arora2025calmqa}. Search-engine--based resources such as NQ \cite{kwiatkowski-etal-2019-natural}, 
support scalable collection of real-world queries, but largely focus on dominant languages. \textbf{\nqaf{} addresses this gap with a unified framework for scalable, multimodal, multilingual, and location-specific data collection}, enabling large-scale curation of culturally grounded, native-language QA.

\section{Conclusions and Future Work}
\label{sec:conclusions}
In this study, we systematize and extend the previously proposed \textit{NativQA} pipeline into a scalable, language-independent framework for building culturally and regionally aligned QA datasets in native languages. It leverages search engines to collect QA pairs from seed queries that native users ask, enabling rapid collection at a chosen locale. The framework has already been used in published studies, and we further evaluate it on 30 additional locations across countries, languages and modalities. Its \textbf{multimodal support} enables new research directions, and caching reduces search-engine API costs. In future, we aim to extend \nqaf{} to more underrepresented locations and languages.


\section{Limitations}
\label{sec:limitations}

The proposed \nqaf{} enables the development of datasets with cultural and native information for everyday knowledge. The \nqaf{} currently uses human-in-the-loop processes in different phases such as seed query creation and manual revision of QA pairs. Though it also supports LLMs for seed query generation and QA validation, however, it is important to have a human-in-the-loop setup to verify the LLMs outputs. The framework does not include a built-in annotation platform for manual QA verification. However, any annotation platform (e.g., Pybossa\footnote{\url{https://github.com/Scifabric/pybossa}}) can be adapted for this purpose. In addition, due to the semi-supervised nature of the approach, it has certain limitations. Some domains are authentic (e.g., Google, Facebook); however, the content they host might not be reliable due to the questionable user-generated content.

\section*{Ethics Statement}
\label{sec:ethics}
The proposed \nqaf{} does not involve collecting any personally identifiable information. Therefore, we do not foresee any issues that may lead to potential risks. 

\section*{Broader Impact}
The \nqaf{} framework supports the creation of culturally grounded, multilingual and multimodal question–answer datasets that better reflect the knowledge needs of diverse communities. By enabling scalable data collection across languages, regions, and dialects, it helps reduce the persistent imbalance in current resources, which remain heavily skewed toward high-resource languages. Overall, \nqaf{} aims to advance the development of more culturally aware LLMs and VLMs.

\bibliography{bib/bibliography}


\setcounter{table}{0}
\setcounter{figure}{0}
\renewcommand{\thetable}{A\arabic{table}}
\renewcommand{\thefigure}{A\arabic{figure}}

\section*{Appendix}
\label{sec:appendix}
\appendix
\section{QA from Search Engines}
\begin{figure}[]
    \centering    
    \includegraphics[width=0.9\columnwidth]{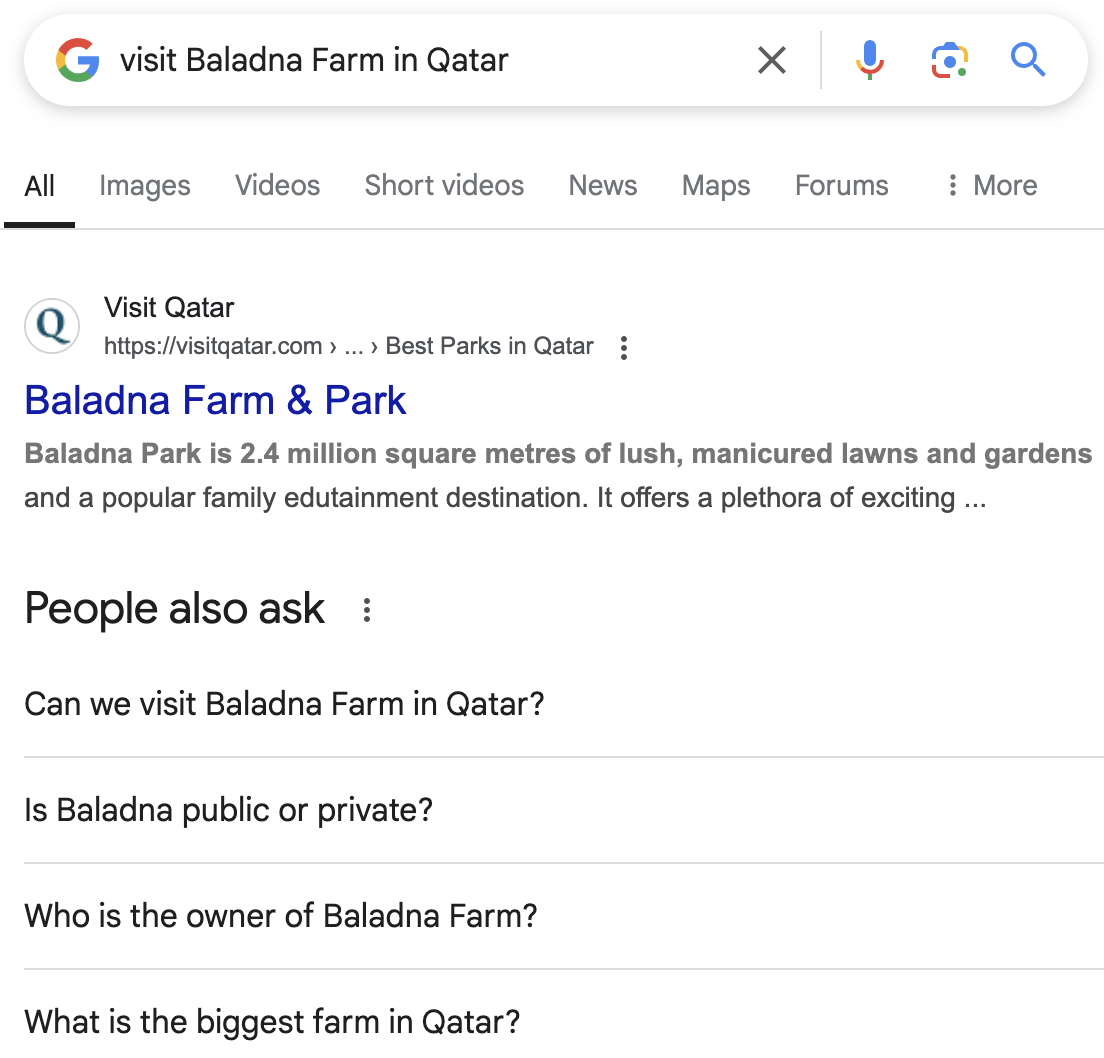}
    \vspace{-0.2cm}
    \caption{QA list in response to a query obtained from Google.}
    \label{fig:google_example}
    \vspace{-0.2cm}
\end{figure}

In Figure \ref{fig:google_example}, we present a screenshot of a Google search results page for the query \textit{``visit Baladna Farm in Qatar.''} The ``People also ask'' section contains related questions, which we used to collect question-answer (QA) pairs. 

\section{Algorithm for QA Collection}
\label{sec:app_algorithm}


\noindent\textbf{Collecting QA Pairs from Search Queries.}
For QA collection we report Algorithm~\ref{alg:collect_qa_pairs}. It outlines a bootstrapped approach for collecting question-answer (QA) pairs using an initial set of seed queries $Q_0$. The process is iterative, aiming to expand the query set and extract more QA pairs at each step. The algorithm takes as input a predefined set of seed queries $Q_0$ and the number of iterations $N_{\text{iter}}$.

At each iteration, for every query in the current set \( S\varrho \), we retrieve QA pairs and related queries using the functions \texttt{ExtractQA(*)} and \texttt{ExtractRelatedQueries(*)}, respectively. \texttt{ExtractQA(*)} returns a set of question-answer pairs $P_{QA}^i$ with corresponding attribution labels \( L \), while \texttt{ExtractRelatedQueries(*)} retrieves additional queries that are semantically or topically related to the input query.

Duplicate QA pairs are removed through the \texttt{DeDuplication(*)} function to ensure the uniqueness of the collected data. The new QA pairs are added to the global set $S_{QA}$, and the related queries are incorporated into the query pool \( S\varrho \) for use in subsequent iterations.


\begin{algorithm}[]
\footnotesize
\caption{\footnotesize Collecting QA pairs using seed queries $\varrho_0$. $P_{QA}^i$: QA pair, $S\varrho_{rel}^i$: related queries. ExtractQA(*) and ExtractRelatedQueries (*) are functions that return questions, $Q$-answers, $A$ pairs with attribution $L$, and related queries, respectively, which are obtained from the search engine for a given query, $q$. DeDuplication (*) removes any duplicate entries from the set to ensure uniqueness.}
\label{alg:collect_qa_pairs}
\begin{algorithmic}[1]
\State \textbf{Input:}
\State \hspace{\algorithmicindent} Seed queries: \( \varrho_0 = \{\hat{\varrho_1}, \hat{\varrho_2}, \ldots, \hat{\varrho_m}\} \)
\State \hspace{\algorithmicindent} Number of iterations: \( N_{iter} \)
\State \textbf{Output:}
\State \hspace{\algorithmicindent} Set of QA pairs: \( S_{QA} \)
\State \hspace{\algorithmicindent} Set of enriched queries: \( S\varrho \)
\State $S_{QA} \gets \emptyset$
\State $S\varrho \gets \varrho_0$
\For {$i$ from 1 to $N_{iter}$}
    \State $P_{QA}^i \gets \emptyset$
    \State $S\varrho_{rel}^i \gets \emptyset$
    \For {$q \in S\varrho$}
        \State $(Q^q, A^q, L^q) \gets \text{ExtractQA}(q)$
        \State $P_{QA}^i \gets P_{QA}^i \cup \{(q', a', l') \mid q' \in Q^q, a' \in A^q, l' \in L^q\}$
        \State $S\varrho_{rel}^i \gets S\varrho_{rel}^i \cup \text{ExtractRelatedQueries}(q)$
    \EndFor
    \State $P_{QA}^i \gets \text{DeDuplication}(P_{QA}^i)$
    \State $S_{QA} \gets S_{QA} \cup P_{QA}^i$
    \State $S\varrho \gets S\varrho \cup S\varrho_{rel}^i$    
\EndFor
\State \textbf{return} $S_{QA}, S\varrho$
\end{algorithmic}
\vspace{-0.1cm}
\end{algorithm}

\section{Distribution of Additional Dataset}
\label{sec:app_additional_dataset}


\begin{table}[t]
\centering
\setlength{\tabcolsep}{1pt} 
\scalebox{0.65}{
\begin{tabular}{clrrrr}
\toprule
\textbf{Language} & \multicolumn{1}{c}{\textbf{Location}} & \multicolumn{1}{c}{\textbf{Train}} & \multicolumn{1}{c}{\textbf{Dev}} & \multicolumn{1}{c}{\textbf{Test}} & \multicolumn{1}{c}{\textbf{Total}} \\ \toprule
\multirow{18}{*}{Arabic} & Abu Dhabi, UAE & 4,896 & 692 & 1,406 & \textbf{6,994} \\
 & Algiers, Algeria & 15,198 & 2,149 & 4,364 & \textbf{21,711} \\
 & Baghdad, Iraq & 17,910 & 2,533 & 5,143 & \textbf{25,586} \\
 & Beirut, Lebanon & 17,748 & 2,510 & 5,096 & \textbf{25,354} \\
 & Cairo, Egypt & 8,631 & 1,221 & 2,478 & \textbf{12,330} \\
 & Damascus, Syria & 7,902 & 1,117 & 2,269 & \textbf{11,288} \\
 & Doha, Qatar & 3,649 & 492 & 988 & \textbf{5,129} \\
 & Gaza, Palestinian & 3,975 & 562 & 1,142 & \textbf{5,679} \\
 & Khartoum, Sudan & 3303 & 467 & 948 & \textbf{4,718} \\
 & Kuwait City, Kuwait & 21,676 & 3,066 & 6,224 & \textbf{30,966} \\
 & Manama, Bahrain & 17,599 & 2,489 & 5,053 & \textbf{25,141} \\
 & Muscat, Oman & 3,479 & 492 & 999 & \textbf{4,970} \\
 & Nouakchott, Mauritania & 13,520 & 1,912 & 3,882 & \textbf{19,314} \\
 & Rabat, Morocco & 17,341 & 2,453 & 4,979 & \textbf{24,773} \\
 & Riyadh, Saudi Arabia & 4,150 & 587 & 1,191 & \textbf{5,928} \\
 & Sanaa, Yemen & 3,373 & 477 & 968 & \textbf{4,818} \\
 & Tripoli, Libya & 3,300 & 467 & 947 & \textbf{4,714} \\
 & Tunis, Tunisia & 10,352 & 1,464 & 2,972 & \textbf{14,788} \\ \midrule
Assamese & Assam, India & 1,131 & 157 & 545 & \textbf{1,833} \\ \midrule
\multirow{2}{*}{Bangla} & Dhaka, Bangladesh & 7,018 & 953 & 1,521 & \textbf{9,492} \\
 & Kolkata, India & 6,891 & 930 & 2,146 & \textbf{9,967} \\ \midrule
\multirow{17}{*}{English} & California, USA & 1,065 & 151 & 306 & \textbf{1,522} \\
 & Dhaka, Bangladesh & 4,761 & 656 & 1,113 & \textbf{6,530} \\
 & Doha, Qatar & 8,212 & 1,164 & 2,322 & \textbf{11,698} \\
 & Florida, USA & 777 & 110 & 223 & \textbf{1,110} \\
 & Georgia, USA & 816 & 115 & 234 & \textbf{1,165} \\
 & Hawaii, USA & 788 & 112 & 226 & \textbf{1,126} \\
 & Illinois, USA & 877 & 124 & 252 & \textbf{1,253} \\
 & Massachusetts, USA & 842 & 119 & 242 & \textbf{1,203} \\
 & Michigan, USA & 796 & 113 & 228 & \textbf{1,137} \\
 & New York, USA & 4,518 & 623 & 1,313 & \textbf{6,454} \\
 & North Carolina, USA & 809 & 115 & 232 & \textbf{1,156} \\
 & Ohio, USA & 862 & 122 & 247 & \textbf{1,231} \\
 & Ontario, Canada & 743 & 105 & 213 & \textbf{1,061} \\
 & Pennsylvania, USA & 786 & 111 & 226 & \textbf{1,123} \\
 & Quebec, Canada & 720 & 102 & 207 & \textbf{1,029} \\
 & Texas, USA & 847 & 120 & 243 & \textbf{1,210} \\
 & Washington, USA & 874 & 124 & 251 & \textbf{1,249} \\ \midrule
Hindi & Delhi, India & 9,288 & 1,286 & 2,745 & \textbf{13,319} \\ \midrule
Nepali & Kathmandu, Nepal & -- & -- & 561 & \textbf{561} \\ \midrule
Turkish & Istanbul, Turkey & 3,527 & 483 & 1,218 & \textbf{5,228} \\ \midrule
\textbf{Total} &  & \textbf{234,950} & \textbf{33,045} & \textbf{67,863} & \textbf{335,858} \\ \bottomrule
\end{tabular}
}
\vspace{-0.2cm}
\caption{Statistics of the data collected using NativQA Framework. 
}
\label{tab:data_split}
\vspace{-0.3cm}
\end{table}

In Table \ref{tab:data_split}, we show the distribution of the dataset across major cities and languages. 

We split the data into training (70\%), development (10\%), and test (20\%) sets.





\section{Annotation Guideline} 
\label{sec:app_annotation_guideline}

\textbf{Manual QA validation - Selecting Best Answer}
\label{app:qa_validation}

The purpose of this task is to evaluate two possible answers for a given question and select the one that is more accurate, clear, and well-structured.
Each question will have two answer options, and the task is to choose the better answer based on the evaluation criteria discussed below.

\noindent
\textbf{Annotation Task}
For each question, there are two answer options. The task is to carefully compare them and select the one that best meets the below criteria:

\begin{enumerate}[label=\arabic*.]
  \item \textbf{Accuracy.}
  The answer must be factually correct and free from misinformation. It should properly address the question and not introduce unrelated details.

  \item \textbf{Clarity \& Readability.}
  The answer should be grammatically correct, well-structured, and easy to read. It should avoid ambiguity, repetition, or fragmented sentences. The information should be presented in a logical and coherent way.

  \item \textbf{Completeness \& Relevance.}
  The answer should provide enough relevant information to fully respond to the question. It should avoid unnecessary details that do not contribute to answering the question. If an answer is too short or vague, it may be less useful than a more complete response.
\end{enumerate}







\noindent
\textbf{Annotation Instructions}
\begin{itemize}
  \item The question and both answer options should be read carefully.
  \item Both answers should be evaluated based on the following three main criteria:
  \begin{itemize}
    \item Accuracy (Is the information correct?)
    \item Clarity (Is it well-written and easy to understand?)
    \item Completeness \& Relevance (Is the question fully and appropriately answered?)
  \end{itemize}
  \item The better answer should be selected based on these criteria.
  \item If the answers are of equal quality, the one that is more precise and easier to understand should be chosen.
  \item If neither answer is acceptable (e.g., both contain major errors), ``neither'' should be selected, and a comment explaining the reasoning should be provided.
\end{itemize}

\noindent
\textbf{Special Cases and Edge Cases}
\begin{itemize}
  \item If one answer is factually correct but lacks clarity, while the other is well-written but factually incorrect, the correct answer should be preferred and revised for clarity if necessary.
  \item If both answers are equally valid in terms of content, the answer that is clearer and better structured should be selected.
\end{itemize}


\noindent
\textbf{Manual QA validation - Answer Scoring.}
\label{app:verifying_quality}
The goal of this task is to assess and, if necessary, revise LLM-edited answers using the information provided in the associated source URLs. Each question includes an LLM-generated answer and its corresponding source. Annotators are expected to verify the quality of the answer based on the source content, and evaluate it according to the criteria outlined below.


\noindent
\textbf{Annotation Task}
We follow the same annotation guideline described by \citet{hasan2024nativqa} to edit the LLM-edited answers.

\noindent
\textbf{Annotation Agreement.}
\label{ssec_annotation_agreement}
In Table~\ref{tab:annotator_likert_score}, we report annotation agreement for a sample of QA pairs. Human annotators rated multiple quality dimensions on a 1–5 Likert scale. Higher Likert scores and higher $r^{*}_{wg(j)}$ values indicate better dataset quality.

\begin{table}[!ht]
    \centering
    \setlength{\tabcolsep}{1pt} 
    \scalebox{0.62}{
    \begin{tabular}{lrrrr}
    \toprule
    \textbf{Location} & \textbf{Clarity} & \textbf{Faithfulness} & \textbf{Informativeness} & \textbf{Plausibility} \\ \midrule
    \multicolumn{5}{c}{\textbf{Average Likert score}} \\ \midrule
    Khartoum, Sudan & 4.41 & 4.42 & 4.26 & 4.29 \\ \midrule
    Sanaa, Yemen & 4.05 & 4.05 & 3.98 & 4.06 \\\midrule
    \multicolumn{5}{c}{\textbf{Agreement index} $r^*_{wg(j)}$} \\ \midrule
    Khartoum, Sudan & 0.85 & 0.85 & 0.85 & 0.85 \\ \midrule
    Sanaa, Yemen & 0.88 & 0.86 & 0.89 & 0.88 \\     
    \bottomrule
    \end{tabular}
    }
    \caption{\textit{Average Likert scale} and \textit{agreement} scores for each human evaluation metric for LLM-edited QA.}
    \label{tab:annotator_likert_score}
\end{table}

\section{Release}
\label{apndix:release}
The \nqaf{} is released under the CC BY-NC-SA 4.0 -- Creative Commons Attribution 4.0 International License: \url{https://creativecommons.org/licenses/by-nc-sa/4.0/}.

\end{document}